%% file: acl_latex.tex
\pdfoutput=1

\documentclass[11pt]{article}

\usepackage[preprint]{acl}

\usepackage{times}
\usepackage{latexsym}

\usepackage[T1]{fontenc}

\usepackage[utf8]{inputenc}

\usepackage{microtype}

\usepackage{inconsolata}

\usepackage{graphicx}
\usepackage{graphicx}
\usepackage{amssymb}
\usepackage{booktabs}
\usepackage{amsthm, amscd, amsfonts, multirow, array}
\usepackage{makecell}
\usepackage{IEEEtrantools}
\usepackage{amsfonts}
\usepackage{amsmath}
\usepackage{enumitem}
\usepackage{colortbl}
\usepackage{textcomp}
\usepackage{makecell}
\usepackage{tabularx}
\usepackage{fontawesome}
\usepackage[textsize=tiny]{todonotes}
\usepackage[capitalize,noabbrev]{cleveref}
\newcommand{\addaddr}[1]{#1}
\newcommand*{\affaddr}[1]{#1} 
\newcommand*{\affmark}[1][*]{\textsuperscript{#1}}
\newcommand{\email}[1]{\href{mailto:#1}{\texttt{#1}}}

\newlength\savewidth

\title{The Role of Visual Modality in Multimodal Mathematical Reasoning: Challenges and Insights}

\author{
Yufang Liu\textsuperscript{\rm 1}\thanks{~~Equal contribution.}\thanks{Work done during an internship at Meituan.}, Yao Du\textsuperscript{\rm 2}\footnotemark[1], Tao Ji\textsuperscript{\rm 3,4}, Jianing Wang\textsuperscript{\rm 2}, \\
\textbf{Yang Liu}\textsuperscript{\rm 2}, 
\textbf{Yuanbin Wu}\textsuperscript{\rm 1}, \textbf{Aimin Zhou}\textsuperscript{\rm 1}, \textbf{Mengdi Zhang}\textsuperscript{\rm 2}, \textbf{Xunliang Cai}\textsuperscript{\rm 2}\\[4pt]
\affaddr{\affmark[1]School of Computer Science and Technology, East China Normal University} \\
\addaddr{\affmark[2] Meituan Inc.} \affaddr{\affmark[3]Fudan University} \affaddr{\affmark[4] Pazhou Laboratory}  \\
\email{yfliu.antnlp@gmail.com}\quad\email{taoji@fudan.edu.cn}\quad\email{ybwu@cs.ecnu.edu.cn} \\
}

\begin{document}
\maketitle
\begin{abstract}
Recent research has increasingly focused on multimodal mathematical reasoning, particularly emphasizing the creation of relevant datasets and benchmarks. Despite this, the role of visual information in reasoning has been underexplored. 
Our findings show that existing multimodal mathematical models minimally leverage visual information, and model performance remains largely unaffected by changes to or removal of images in the dataset. We attribute this to the dominance of textual information and answer options that inadvertently guide the model to correct answers. To improve evaluation methods, we introduce the HC-M3D dataset, specifically designed to require image reliance for problem-solving and to challenge models with similar, yet distinct, images that change the correct answer.
In testing leading models, their failure to detect these subtle visual differences suggests limitations in current visual perception capabilities.
Additionally, we observe that the common approach of improving general VQA capabilities by combining various types of image encoders does not contribute to math reasoning performance. This finding also presents a challenge to enhancing visual reliance during math reasoning. Our benchmark and code would be available at
\href{https://github.com/Yufang-Liu/visual_modality_role}{https://github.com/Yufang-Liu/visual\_modality\_role}.

\end{abstract}

\input{introduction}
\input{related_work}

\input{experiment}

\section{Conclusion}
We study the importance of the visual modality within multimodal mathematical models. Our findings indicate that the performance of existing multimodal mathematical models does not significantly deteriorate when images are shuffled or even removed. We propose HC-M3D benchmark to measure the model's sensitivity to changes in images. Experiments demonstrate that the model does not alter its answers corresponding to changes in the images when there is a high relevance between images and answers. We further discover that enhancing the dependency on the visual modality in the mathematical domain presents significant challenges, and we leave further exploration to subsequent work.

\section*{Limitations}
Firstly, this paper focuses on the importance of the visual modality in mathematical LVLMs during pre-training and instruction fine-tuning, without addressing reasoning stages such as chain-of-thought, which will be part of our future work. Secondly, due to the high cost and time-consuming nature of manual annotation, especially generating new images for different options of the same question, the scale of our proposed HC-M3D dataset is relatively small, consisting of only 1,851 samples. Lastly, although the method of integrating multiple visual experts shows performance improvements in general VQA tasks, it offers little benefit (and in some cases, even negative impact) when applied to mathematical reasoning evaluation. This highlights the differences between mathematical reasoning tasks and general tasks, urging us to explore solutions tailored to mathematical reasoning tasks.

\section*{Ethics Statement}

The HC-M3D dataset is sourced from publicly available academic datasets or manually collected math problems. The annotations were performed by a professional labeling team from the company, with the task of filtering math problems and generating images. Therefore, no ethical issues are involved.

\section*{Acknowledgement}
The authors wish to thank all reviewers for their
helpful comments and suggestions. The corresponding authors are Tao Ji, Yuanbin Wu, Aimin
Zhou and Mengdi Zhang. 
The computations in this research were performed using the CFFF platform of Fudan University.

\bibliography{custom}

\appendix

\input{appendix}

\end{document}

%% file: introduction.tex
\section{Introduction}

Recent advancements in Large Vision-Language Models (LVLMs)~\cite{DBLP:journals/corr/abs-2304-00685} have demonstrated remarkable potential in tasks that require the seamless integration of visual and linguistic understanding, such as image captioning~\cite{DBLP:conf/eccv/LinMBHPRDZ14}, visual question answering~\cite{DBLP:conf/iccv/AntolALMBZP15}, visual grounding~\cite{DBLP:conf/eccv/YuPYBB16}, and autonomous agents~\cite{DBLP:journals/corr/abs-2309-07864, DBLP:journals/corr/abs-2401-03568}. Among these, the domain of multimodal mathematical reasoning~\cite{DBLP:journals/corr/abs-2402-14804, DBLP:conf/iclr/LuBX0LH0CG024, DBLP:conf/eccv/ZhangJZLGQZLCQGL24} emerges as a particularly challenging and pivotal application. This task requires models to accurately interpret and abstractly represent mathematical concepts, such as geometry, through a tight collaboration between textual descriptions and visual elements in diagrams. The ability to align these modalities and resolve complex problems lies at the heart of LVLMs' capabilities, positioning multimodal mathematical reasoning as a critical field for advancing their effectiveness in abstract and logic-driven applications.

\begin{figure}[t]
    \centering
    \includegraphics[scale=0.35]{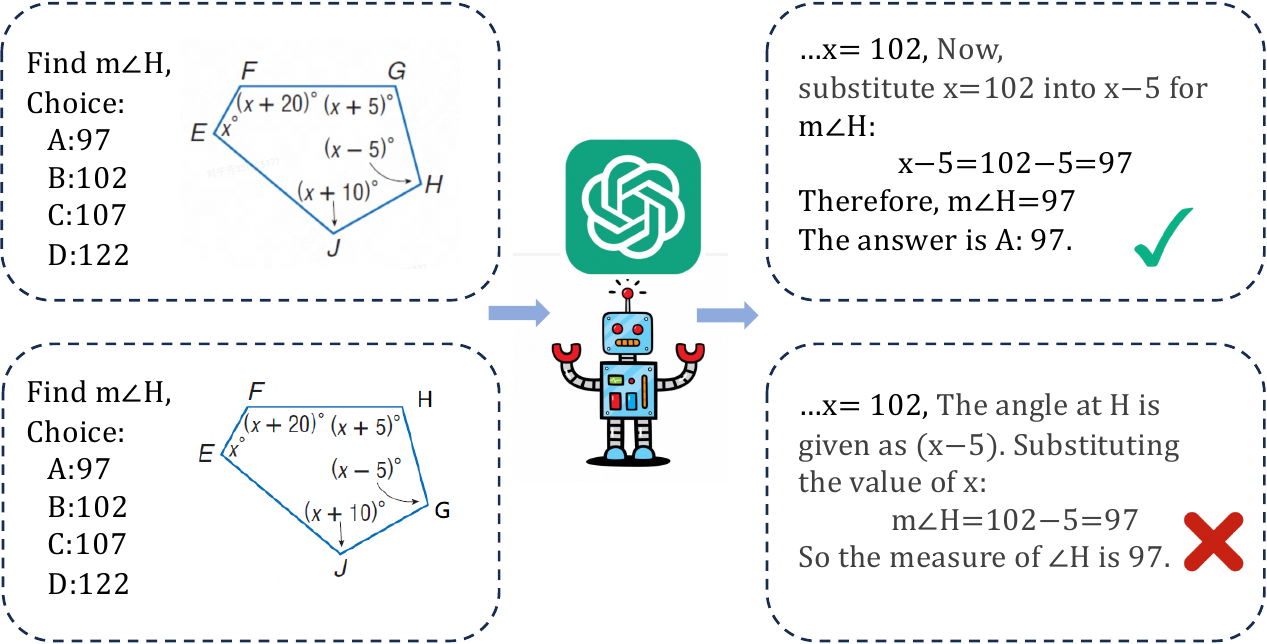}
    \caption{An illustrative example of GPT-4o on the HC-M3D. When modifying the image (by swapping the positions of points G and H) while keeping the original question unchanged, the model's output failed to correspondingly adjust, resulting in an incorrect response.}
    \label{fig:demo}
\end{figure}

There are three common approaches to enhancing mathematical reasoning capabilities in LVLMs: prompting, fine-tuning, and chain-of-thought (CoT) reasoning. 
The prompting approach~\cite{DBLP:conf/iclr/LuBX0LH0CG024,DBLP:journals/corr/abs-2402-17971} seeks to unlock the potential of LVLMs through carefully designed prompts. 
Fine-tuning~\cite{DBLP:conf/emnlp/ShiHBL0NBL24, DBLP:journals/corr/abs-2312-11370}, on the other hand, focuses on constructing higher-quality and more diverse datasets to enhance model performance. 
Chain-of-thought~\cite{DBLP:journals/corr/abs-2406-09403}reasoning mimics human reasoning by generating detailed step-by-step thought processes to tackle complex mathematical tasks. 
These methods rely heavily on training data tailored for mathematical understanding. 
Existing research primarily emphasizes improving the diversity of problem types~\cite{DBLP:journals/corr/abs-2312-11370}, refining the conciseness and clarity of questions~\cite{DBLP:journals/corr/abs-2407-08739}, and providing more detailed answer explanations~\cite{DBLP:journals/corr/abs-2409-00147}, with model performance evaluated on various benchmarks. 
\textit{However, current datasets often fail to establish a strong connection between mathematical images and text.}

In this work, we explore whether images are genuinely learned and utilized by models for mathematical reasoning within current multimodal mathematical methods and examines the role they play in the process.
We observe that in existing multimodal math models, the benefit brought by image patterns is minimal. 
Even if the images in the training set are shuffled or even removed, the impact on model performance is not significant (On some evaluation datasets, perturbing the visual modality even resulted in higher scores). 
By analyzing existing evaluation methods, we suspect that there are two main reasons why the role of visual patterns in multimodal math reasoning has been overestimated: first, the textual information is overly rich, and the model can make correct predictions based on text alone; second, the options leak the answers, and the model can guess the correct answer based on the options. 

Hence, we propose the HC-M3D dataset, a \textbf{h}uman-\textbf{c}rafted \textbf{m}uli\textbf{m}odal \textbf{m}athematical \textbf{d}ataset, comprising 1,851 samples meticulously selected by humans, that ensures that the questions depend on the images and provides an additional image that looks similar but changes the correct answer for over 400 problems. 
We observe whether models can identify these subtle differences in the images and make correct predictions. In evaluating existing leading models, we find that they fail to accurately identify these differences, and in more than half of the cases, they stick to their original predictions. Figure~\ref{fig:demo} presents a representative example.

Furthermore, we attempt to delve into the causes of this phenomenon. Considering that the model's predictions remain unchanged after image alterations, we infer that the image encoder (mainly the CLIP encoder) fails to effectively recognize such subtle differences, a phenomenon also observed in other works~\cite{DBLP:conf/emnlp/LiuJSWZ24, DBLP:conf/cvpr/Tong0Z0LX24}. 
However, we observe that the common approach of improving general VQA capabilities by combining various types of image encoders does not contribute to math reasoning performance. This finding also presents a challenge to enhancing visual reliance during math reasoning.


Our contributions can be summarized as follows:
\begin{itemize}[noitemsep,leftmargin=*,topsep=0pt]
    \item We show that image shuffling or removal has minimal impact on performance, revealing an overestimation of these models' reliance on visual input. 
    \item We introduce the HC-M3D dataset with 1,851 human-annotated samples, ensuring image-dependent questions. For 429 questions, similar but altered images are included, yet most mathematical LVLMs fail to adjust predictions, often remaining unchanged.
    \item We demonstrate the challenge of enhancing visual reliance in mathematical reasoning, as combining image encoders proves ineffective.
\end{itemize}

Based on these observations, we suggest several potential directions for future research: constructing higher-quality datasets with a stronger reliance on visual data (e.g., in the image-caption pre-training task, describing the differences between two images rather than generating captions for a single image.), improving image encoders to capture more fine-grained mathematical information, and designing better loss functions to enhance the model's dependence on the visual modality.

%% file: related_work.tex
\section{Related Work}


Mathematical reasoning has garnered increasing attention from researchers and has become a critical benchmark for evaluating LLMs. 
Specialized LLMs such as MetaMath~\cite{DBLP:conf/iclr/YuJSYLZKLWL24}, Math-Shepherd~\cite{DBLP:conf/acl/WangLSXDLCWS24}, WizardMath~\cite{DBLP:journals/corr/abs-2308-09583}, and DeepSeekMath~\cite{DBLP:journals/corr/abs-2402-03300} have been developed to tackle these tasks. 
One of the challenges in multimodal mathematical reasoning lies in visual-textual reasoning, involving geometric diagrams, scientific charts, and function graphs. 
Due to the scarcity of training data, existing methods often employ the LLaVA architecture for training, aiming to generate higher-quality and more diverse  training data. For instance, G-LLaVA~\cite{DBLP:journals/corr/abs-2312-11370} obtains image caption data for alignment by rewriting the original QA data and paraphrases the remaining QA data to augment the SFT data. Math-LLaVA~\cite{DBLP:conf/emnlp/ShiHBL0NBL24} synthesizes a large number of QA pairs based on seed questions, while MAVIS~\cite{DBLP:journals/corr/abs-2407-08739} automatically constructs QA pairs through templates.

With the rapid advancement of LVLMs, there is a growing need for high-quality benchmark tests to assess math problem-solving in visual settings. While previous attempts like GeoQA~\cite{DBLP:conf/acl/ChenTQLLXL21} and Geometry3K~\cite{DBLP:conf/acl/LuGJQHLZ20} primarily focused on geometric problems, the recently proposed MathVista~\cite{DBLP:conf/iclr/LuBX0LH0CG024} takes it a step further by encompassing a diverse range of multimodal tasks involving mathematical reasoning. Mathverse~\cite{DBLP:conf/eccv/ZhangJZLGQZLCQGL24}, on the other hand, emphasizes textual and visual richness to gauge whether models truly understand image content.

%% file: experiment.tex
\section{Does Visual Modality Matter for Mathematical LVLMs?}\label{sec:visual_role}

To validate the importance of visual modality, we replicated the training processes of mainstream mathematical LVLMs under a unified architecture, with the primary differences lying in the training sets. By perturbing the alignment between images and text—either by shuffling correct pairs or masking image information—we observed that the data utilized in mathematical LVLMs fails to enable the model to establish a strong correlation between images and text.

\subsection{Setups}

\paragraph{Mathematical LVLMs}
We study the recently proposed mathematical LVLMs, including:
\textbf{G-LLaVA}~\citeyearpar{DBLP:journals/corr/abs-2312-11370} which constructs an enhanced multimodal geometric dataset, Geo170K, based on the scalability of geometric problems and fine-tunes it based on LLaVA. 
Similarly, \textbf{MathLLaVA}~\citeyearpar{DBLP:conf/emnlp/ShiHBL0NBL24} also establishes a large-scale multi-type mathematics problem dataset, which includes filtered image clarity and question complexity for Chart Question Answering, Geometric Problem Solving, Math Word Problems, Textbook Question Answering, and Visual Question Answering. 
\textbf{MAVIS}~\citeyearpar{DBLP:journals/corr/abs-2407-08739} utilizes an automatic data engine to generate mathematical visual datasets covering areas such as plane geometry, analytic geometry, and functions, effectively bypassing the high cost of manual annotation. 
\textbf{MultiMath}~\citeyearpar{DBLP:journals/corr/abs-2409-00147} collects textbooks, exercises, and exam questions from the K-12 curriculum and employs GPT to generate detailed CoT processes, featuring bilingual data.

\input{lvlm_eval}

\paragraph{Reproduction Details}
Since all LVLMs are based on the LLaVA architecture, which connects the language model core to the visual encoder, we standardize and choose \texttt{deepseek-math-rl-7B} as language model and \texttt{clip-vit-large-patch14-336}  as image encoder for reproduction (same as~\citet{DBLP:journals/corr/abs-2409-00147}). 
The training process is divided into three stages: the pre-training stage, where the connection module of LLaVA is fine-tuned on alignment data (including math alignment data, if available); the SFT stage, where the alignment module and the language model are fine-tuned on LLaVA's general data; and the Math SFT stage, where the alignment module and the language model are further fine-tuned on multimodal math QA questions.
Detailed statistcs can be found in Table \ref{tab:training_details}.

\paragraph{Mathematical Benchmarks $\mathcal{D}_{\text{math}}$}
We evaluate the following datasets: 
\textbf{MathVerse}~\citeyearpar{DBLP:conf/eccv/ZhangJZLGQZLCQGL24} offers a collection of 3,940 high-quality mathematical problems across various disciplines, enriched with relevant charts, showcasing varied dependencies on visual and textual cues. 
Conversely, \textbf{MathVista}~\citeyearpar{DBLP:conf/iclr/LuBX0LH0CG024} presents a comprehensive benchmark with 1,000 instances from 28 multimodal datasets, emphasizing their acute visual comprehension and intricate reasoning skills.
\textbf{GeoQA}~\citeyearpar{DBLP:conf/acl/ChenTQLLXL21} consists of 754 pieces of data, mainly multiple-choice questions about plane geometry.
\textbf{MATHVision}~\citeyearpar{DBLP:journals/corr/abs-2402-14804} comprises 304 high-quality mathematical problems sourced from authentic mathematics competitions. The questions span 16 distinct mathematical disciplines and are categorized into 5 difficulty levels.
\textbf{WeMath}~\citeyearpar{DBLP:journals/corr/abs-2407-01284} consists of 1,740 visual math questions, covering 67 hierarchical knowledge concepts and 5 levels of knowledge granularity. \footnote{For Mathverse, MathVita, and MathVision, as they contain multiple test sets, we select the ``testmini'' subset for evaluation.}

\input{text_model}

\subsection{Visual Modality Perturbation}

\paragraph{Data w/ correct images \faFileImageO~\faLongArrowRight~\faFileTextO:} The standard training process, the model can learn the correspondence between images and text as well as the templates for question-answering during the Math SFT stage.

\paragraph{Data w/ random images \faFileImageO~\faRandom~\faFileTextO:} During the math SFT stage, by shuffling the correspondence between images and texts while keeping the picture distribution unchanged, we ensure that after the correspondence is disrupted, the Q\&A for the original picture will be assigned to another picture. In this process, the model can learn the statistical information of the picture distribution and the template for answering questions.

\paragraph{No images \faFileO~\faLongArrowRight~\faFileTextO:} For the Math SFT stage, we remove the images from the training set and retain the question and answer. This helps the model learn the template for question-answering and enhances the quality of the responses.

\paragraph{Observations} The results are presented in the \Cref{table:visual_modality}, we can observe that:
\begin{itemize}[leftmargin=*]
    \item Firstly, after standardizing the model structure and training methodologies, the discrepancies among different approaches diminish (compared to the contrasts presented in~\citet{DBLP:journals/corr/abs-2407-08739, DBLP:journals/corr/abs-2409-00147}). Contrary to the direct fintuning from LLaVA for alignment as discussed in the G-LLaVA paper~\cite{DBLP:journals/corr/abs-2312-11370}, we notice a significant improvement with a three-stage training approach, aligning with the findings of ~\citet{DBLP:journals/corr/abs-2409-00147}. The MultiMath model exhibits superior performance across datasets, with the exception of MathVerse, attributable to its diverse dataset and detailed reasoning processes.
    \item Secondly, we observe that substituting the correct images with shuffled ones only slightly impacts model performance. This trend is consistent across different models and datasets. Specifically, average performance decreases by 0-4 percentage points. In contrast, the GeoQA and MathVisions datasets may even experience some degree of improvement. From a model perspective, high-performance models are more affected by image shuffling than lower-performance models. Compared to other models, the G-LLaVA model exhibits less performance fluctuation.
    \item Lastly, similar experimental outcomes are observed in the absence of images. These results suggest that during the mathematical SFT phase, image patterns do not significantly enhance reasoning performance, indicating that models primarily rely on text to learn question-and-answer capabilities in the mathematical domain.
\end{itemize}


\paragraph{Compared with general tasks $\mathcal{D}_{\text{general}}$} 
To observe if a similar phenomenon exists in the general VQA task, we follow the training process of LLaVA-1.5 with vicuna-7B~\cite{vicuna2023} and examine the impact of visual information on VQA models by comparing the performance differences between three scenarios: shuffled images, removed images, and provided correct images. We select the following commonly used VQA evaluation datasets: VQAv2~\citeyearpar{DBLP:conf/cvpr/GoyalKSBP17}, MMBench~\citeyearpar{DBLP:conf/eccv/LiuDZLZZYWHLCL24}, MM-Vet~\citeyearpar{DBLP:conf/icml/YuYLWL0WW24}, and MME~\citeyearpar{DBLP:journals/corr/abs-2306-13394}.

The results in Table~\ref{table:visual_modality} indicate a significant impact on model performance in general VQA tasks when images are either shuffled randomly or not provided at all. 
Not providing images leads to markedly better performance than offering random images: without images, model performance drops by approximately 23\% on VQAv2 and 19\% on MMBench; with random images, the decline is more drastic at 42\% and 61\%, respectively. This suggests that models can detect inconsistencies between randomized image content and text, thereby disrupting learning. In contrast, the trend of performance decline in the mathematical domain is much weaker. We speculate this is because multimodal mathematical reasoning tasks rely more on generative aspects and less on image comprehension, a discrepancy that may not be as apparent in general VQA tasks.

\section{Issues in Existing Benchmarks}
We further explore the reasons for the overestimation of the image modality in multi-modal mathematical reasoning tasks. Through a careful examination of existing evaluation sets, we identify two main issues.

Firstly, a large proportion of the existing evaluation dataset can be answered correctly based on the text alone, which does not accurately reflect whether the image modality is effectively utilized. 
We conduct an evaluation on 7 common textual models and and 3 vision language models.
The experimental results are shown in Table \ref{tab:text_model}. As compared to the results in Table \ref{table:visual_modality}, the existing multimodal math methods show limited improvement based on the language model \texttt{DeepSeek-Math-7B-RL}. For example, MultiMath exhibits enhanced performance on the MathVerse and Wemath datasets, increasing from 20.7 to 29.6 and from 33.3 to 44.7, respectively. However, it even experiences a degree of decline on the MathVision dataset, dropping from 11.5 to 7.2.

We also observe that as the capability of the language model improves, the performance of the text model is further enhanced. In terms of average performance across all datasets, when the language model is switched from \texttt{DeepSeek-Math-7B-RL} to \texttt{QWen-2.5-Math-7B-Instruct}, the score increase from 26.3 to 33.3. Meanwhile, the average performance of the G-LLaVA model is only 35.2, indicating that the performance of language models is very close to that of multimodal models.

Secondly, the options in the question may leak the answer, and the model can guess the answer based on the option information. We shuffle the order of the options in the multiple-choice questions, observe the changes in the model's predictions before and after, and determine whether the model truly understands the question or is just making predictions based on option information.
Table \ref{tab:shuffle} shows the experimental results. We calculate the original accuracy (CR: Correct), the proportion of predictions are correct both before and after shuffled (BC: Both Correct), and the proportion of consistent predictions before and after (AR: Agreement). As can be seen from the table, in the case of multiple-choice questions, the BC indicator is significantly lower than the CR indicator. For example, the BC/CR indicators of the MultiMath model on the three datasets are 9.5\%, 76.1\%, and 80.7\%, respectively. This indicates that the model's single-performance is not sufficient to reflect its overall performance on the dataset. Additionally, a very low BC/CR could indicate potential quality issues with the dataset.

\begin{table}[t]
\centering
\small
\setlength\tabcolsep{3pt}
\resizebox{\linewidth}{!}{
\begin{tabular}{lrrrrrrrrr}
\toprule
\multirow{2}{*}{\textbf{Model}} & \multicolumn{3}{c}{\textbf{GeoQA}} & \multicolumn{3}{c}{\textbf{MathVerse}} & \multicolumn{3}{c}{\textbf{MathVista}}           \\
\cmidrule(r){2-4} \cmidrule(r){5-7} \cmidrule(r){8-10}
& \textbf{CR}$_\uparrow$ & \textbf{BC}$_\uparrow$ & \textbf{AG}$_\uparrow$ & \textbf{CR}$_\uparrow$ & \textbf{BC}$_\uparrow$& \textbf{AG}$_\uparrow$& \textbf{CR}$_\uparrow$& \textbf{BC}$_\uparrow$ & \textbf{AG}$_\uparrow$\\
\midrule
G$_{\text{LLaVA}}$    & 68.0 & \textbf{16.0} & \textbf{20.8} & 44.4 & 30.3 & 50.6 & 51.7 & 40.9 & 70.4 \\
Math$_{\text{LLaVA}}$ & 57.4 & 7.6  & 13.3 & 40.0 & 26.2 & 53.4 & \textbf{58.7} & \textbf{49.3} & \textbf{75.4} \\
MAVIS     & 70.6 & 8.2  & 10.1 & 41.9 & 29.0 & 49.1 & 54.6 & 44.8 & 72.0 \\
MultiMath & \textbf{74.4} & 7.1  & 8.6  & \textbf{48.1} & \textbf{36.6} & \textbf{56.6} & 55.0 & 44.4 & 72.4 \\
\bottomrule
\end{tabular}
}
\caption{Results of shuffling the order of the multiple-choice options in the dataset. The table indicates the accuracy (CR: Correct), predictions are correct both before and after shuffled (BC: Both Correct), and predictions remained consistent across both attempts (AG: Agreement).}
\label{tab:shuffle}
\end{table}

\begin{figure*}[htbp]
    \centering
    \includegraphics[width=0.83\textwidth]{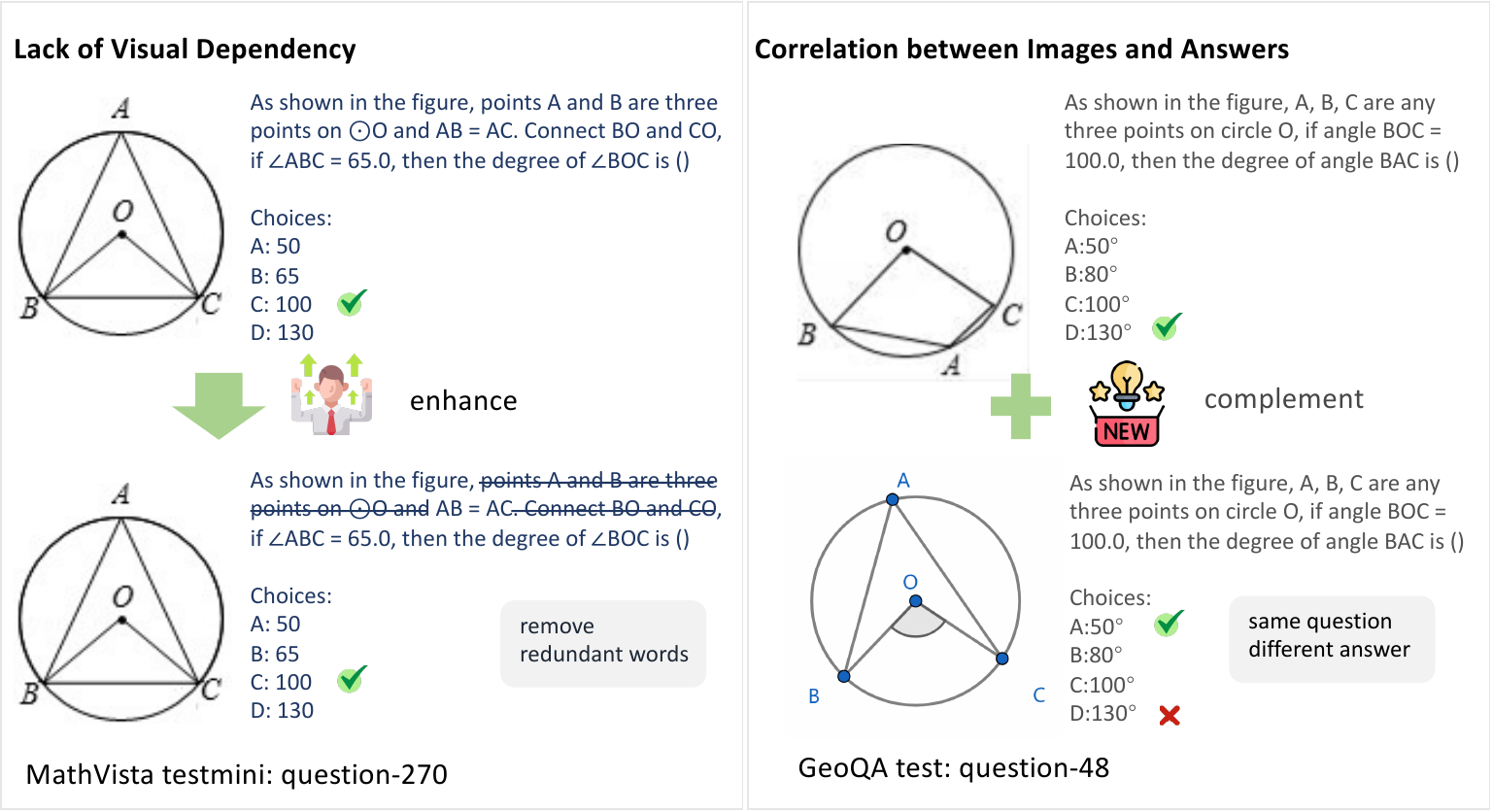}
    \caption{Examples from our constructed HC-M3D benchmark. The left image demonstrates a scenario where the original data lacks visual dependency, allowing for the reconstruction of the question's corresponding image solely from text. This is addressed by rewriting or possibly changing the image to ensure the question requires image dependency for a correct answer. The right image shows an example where supplementing the original question with a similar image alters the correct answer. More examples can be found in Figure~\ref{fig:dependency} and Figure~\ref{fig:connection}.}
    \label{fig:dataset}
\end{figure*}

Based on our observations, we believe that existing multimodal math datasets do not accurately measure the importance of the image modality. Since a large portion of the questions can be answered by textual models alone and the model can guess the answers, these issues lead to an overestimation of the capability of the visual modality, especially in multiple-choice questions. 
To improve the accuracy of the assessment, it is essential to mitigate the above two issues. Establishing a suitable benchmark will significantly advance the research on multimodal mathematic models.

\section{The HC-M3D Benchmark}\label{sec:benchmark}

Based on the analysis above, we propose HC-M3D, a high-quality, vision-dependent multimodal benchmark. 
Our dataset construction adheres to the following three principles. 
\begin{itemize}[leftmargin=*]
    \item First, correctness of the data, meaning that the collected data must be solvable based on the question and image asked and the answer is correct.
    \item Second, visual dependency, indicating that the data must rely on images for accurate answers. 
    \item Lastly, high correlation between images and answers. Where possible, we aim to keep the question unchanged but alter the image to ensure that the answer changes. This methodology tests whether the model can detect changes in the images and make correct predictions.
\end{itemize}
\begin{table}[t]
\centering
\small
 \begin{tabular}{lr}
 \toprule
 \textbf{Statistic} & \textbf{Number} \\
 \midrule
  Total questions & 1,851 \\
  ~- Multiple-choice questions & 1,851 (100.0\%) \\
  ~- \textbf{Newly collected questions} & \bf 1,219 (65.9\%) \\
  ~- \textbf{Newly collected images} & \bf1,084 (58.6\%) \\
 \midrule
 Data Souce &  \\
  ~- GeoQA &  896 (48.4\%) \\
  ~- MathVista & 273 (14.7\%) \\
  ~- Jingyou Net & 684 (37.0\%) \\
  \midrule
  Language & \\
  ~- English & 1,052 (56.8\%)\\
  ~- Chinese & 799 (43.2\%)\\
  \midrule
 Number of unique images & 1,851 (100.0\%) \\
 Number of unique questions & 993 (53.6\%) \\
 \bottomrule
 \end{tabular}
 \caption{Statistical Results on the HC-M3D Dataset.}
 \label{tab:dataset}
\end{table}

\paragraph{Data Curation} Our dataset is sourced from the GeoQA and MathVista GPS subsets, along with a selection of questions from  the Chinese Jingyou Net~\footnote{\href{https://www.jyeoo.com/}{https://www.jyeoo.com/}}, primarily focusing on plane geometry knowledge. 
We manually annotate each sample for correctness and visual dependency, modifying the original questions when these criteria are not met and attempting to supply an additional image for the question to change the correct answer. 
Instructions for human annotators of these two steps are shown in Figure~\ref{fig:instruction1} and Figure~\ref{fig:instruction2}.
Ultimately, we obtain 1,851 samples, including 429 questions for which an additional image was provided.
Statistical results for the dataset are presented in Table \ref{tab:dataset}.

Figure \ref{fig:dataset} illustrates the process of constructing the dataset. Upon manually identifying a lack of visual dependency in the data, the text (or image) is revised to ensure that the question requires reliance on the image for an answer. For instance, the image on the left can be drawn based on the question, demonstrating low dependency on the image; thus, a revision is necessary. Under conditions where dependency is met, as with the image on the right, changing the position of point A on the circle alters the correct answer from D to A.
In this scenario, there is a high correlation between the image and the correct answer, necessitating that the model explicitly perceives the differences in the image to make accurate predictions.

Our HC-M3D stands out by modifying images and using a controlled-variable method to verify whether the model can truly and accurately understand the image content. The challenge with MathVerse lies in precisely distributing information, a process that is somewhat ambiguous. In contrast, H3-M3D changes the answer solely through image modifications, with the text remaining unchanged. Our control over the text information is thorough and successful (annotators simply need to determine whether the modified image corresponds to a new answer).

\paragraph{Metrics}
Performance on the dataset is evaluated based on the following metrics. Accuracy over all data (ALL); accuracy on a subset (DI: Diverse Image) consisting of 858 samples that share similar questions but have different images and answers; accuracy (BC: Both Correct) for cases within the DI subset where both samples sharing the same question are correctly predicted; and consistency rate (AG: Agreement) within the DI subset for samples that share the same question and are predicted consistently. Since these samples have differing answers, a lower AG metric is preferred. \footnote{The visual control experiment results on the proposed HC-M3D benchmark can be found in Table~\ref{table:visual_modality_hcm3d}.}

\begin{table}[t]
\centering
\small
\setlength\tabcolsep{3pt}
\resizebox{\linewidth}{!}{
\begin{tabular}{lccccc}
\toprule
\textbf{Model}        & \bf \#Param. & \textbf{ALL} $\uparrow$ & \textbf{DI} $\uparrow$ & \textbf{BC} $\uparrow$ & \textbf{AG} $\downarrow$ \\
\midrule
G-LLaVA            & 7B  & 45.4 & 41.5 & 15.2 & 52.2 \\
MathLLaVA         & 7B  & 39.6 & 37.2 & 8.4  & 75.3 \\
MAVIS             & 7B  & 42.8 & 37.7 & 9.8  & 58.0 \\
MultiMath         & 7B  & 49.2 & 44.8 & 16.6 & 56.9 \\
InternVL2         & 8B  & 41.9 & 38.3 & 16.6 & \textbf{34.0} \\
QWen2-VL-Instruct & 8B  & 40.9 & 40.7 & 13.8 & 58.7 \\
InternVL2-Llama3  & 76B & 47.4 & 45.5 & 19.4 & 40.8 \\
QWen2-VL-Instruct & 76B & \textbf{51.8} & \textbf{48.3} & \textbf{20.3} & 51.5 \\
GPT-4o             & /   & 49.0 & 45.8 & 19.1 & 42.0  \\
\bottomrule
\end{tabular}
}
\caption{Evaluation results on the HC-M3D benchmark. The $\uparrow$ and $\downarrow$ arrows respectively indicate that higher or lower values are preferred. }
\label{tab:dataset_m3d}
\end{table}

\begin{table*}[t]
\centering
\small
\setlength\tabcolsep{3pt}
\resizebox{\linewidth}{!}{
\begin{tabular}{cccccccccccc}
\toprule
\multirow{2}{*}{\textbf{Image Encoder}} & \multicolumn{5}{c}{\textbf{General VQA Performance after SFT stage}} & \multicolumn{6}{c}{\textbf{Math Performance after Math SFT stage}} \\
\cmidrule(r){2-6} \cmidrule(r){7-12}
& \textbf{VQA$^{v2}$} & \textbf{MMBench} & \textbf{MM-Vet} & \textbf{MME$^P$} & \textbf{MME$^C$} & \textbf{GeoQA} & \textbf{M.Verse} & \textbf{M.Vista} & \textbf{M.Vision} & \textbf{WeM.} & \textbf{HC-M3D} \\
\midrule
clip-B &72.6 &60.7 &25.7 &1365.8 &273.6  &71.0 &23.0 &31.4 &9.5 &38.7 &41.2 \\
clip-L & 77.0 & 64.8 & 30.4 & 1481.0 & 269.6 & 68.0 & \textbf{24.7} & \textbf{34.9} & 8.9 & \textbf{39.3} & \textbf{45.4} \\
\hline
\multicolumn{12}{c}{(a) combine features by concatenating the hidden feature} \\
clip-B+ siglip-B & 74.6 & 62.0 & 25.7 & 1397.5 & 257.5 & \textbf{72.3} & 22.2 & 31.1 & 8.2 & 38.1 & 40.0 \\
clip-L+dinov2-L  & 78.4 & 66.9 & 30.9 & 1500.1 & 253.2 & 70.3 & 23.2 & 30.0 & \textbf{12.2} & 38.6 & 40.3 \\
clip-L+siglip-B+dino-B  & 74.4 & 62.5 & 25.5 & 1404.0 & 280.0 & 70.8 & 22.5 & 31.0 & 8.9 & 36.0 & 40.5 \\
\hline
\multicolumn{12}{c}{(b) combine features by concatenating image tokens} \\
clip-L+siglip-L & 78.7 & 66.6 & 33.0 & 1445.2 & 266.4 & 69.9 & 22.6 & 30.8 & 8.9 & 37.2 & 39.2 \\
clip-L+dinov2-L & 78.6 & 66.7 & 32.3 & 1494.7 & 269.6 & 70.7 & 22.3 & 28.7 & 7.6 & 35.7 & 39.2 \\
clip-L+siglip-L+dinov2-L & \textbf{80.0} & \textbf{69.8} & \textbf{35.8} & \textbf{1524.6} & \textbf{301.8} & 69.9 &  22.6     & 28.5 & 8.9  & 38.1 & 39.4 \\
\bottomrule
\end{tabular}
}
\caption{Performance on general VQA and mathematical tasks. 
clip-L, sigLip-L, and dino-L correspond to \texttt{clip-vit-large-patch14-336}, \texttt{siglip-so400m-patch14-384}, and \texttt{dinov2-large} models,respectively.  
siglip-B (siglip-base-patch16-224) and dino-B (\texttt{dino-vitb16}) are utilized alongside clip-B (\texttt{clip-vit-base-patch16}) for matching image token numbers in combinations.
Detailed results can be found in Table~\ref{tab:detail_mme} and Table~\ref{tab:detail_math}.}
\label{tab:image_encoder}
\end{table*}

\begin{table}[t]
\centering
\resizebox{0.85\linewidth}{!}{
\begin{tabular}{ccc}
\toprule
\textbf{Image Encoder} & \textbf{POPE} & \multicolumn{1}{l}{\textbf{MathPOPE}} \\
\midrule
clip-B & 81.4     &  72.9    \\
clip-L & \textbf{85.2} & 73.7 \\
\hline
\multicolumn{3}{c}{(a) concatenating the hidden feature} \\
clip-B+siglip-B & 82.2 & 72.6 \\
clip-L+dinov2-L  & 84.4 & 70.1 \\
clip-L+siglip-B+dino-B & 80.1 & 76.2\\
\hline
\multicolumn{3}{c}{(b) concatenating image tokens} \\
clip-L+siglip-L & 83.9 & 67.6 \\
clip-L+dinov2-L & 84.2& 71.9 \\
clip-L+siglip-L+dinov2-L & 84.1 & \textbf{81.0} \\      
\bottomrule
\end{tabular}
}
\caption{F1 performance on both POPE and MathPOPE datasets. POPE primarily focuses on questions about objects in images, whereas MathPOPE concentrates on points existing in geometric images.}
\label{tab:pope}
\end{table}

\paragraph{Evaluation}
Experimental results are shown in Table \ref{tab:dataset_m3d}. It is observed that among existing multimodal mathematical models, MutiMath achieves the highest accuracy of 49.2, comparable to the performance of \texttt{GPT-4o}. The open-source model \texttt{QWen2-VL-Instruct-76B}, reaches an highest accuracy of 51.8. 
Furthermore, in the DI subset, which accounts for image and answer relevance, the proportion of question pairs correctly predicted/Performance on the DI subset (BC/DI) remains below 50\%. 
This indicates that the models do not deeply understand questions for which half of the answers are correct; changing the image leads to incorrect predictions. At the same time, the AG metric for different models is generally above 50\%, suggesting that models do not appropriately change their answers when the image is replaced. The \texttt{GPT-4o} and \texttt{InternVL2} models perform best in this metric.

\section{Challenges in Enhancing Visual Dependency}
We observe that existing multimodal mathematical models exhibit weak reliance on visual information~(Section \ref{sec:visual_role}), demonstrating insufficient sensitivity to changes in images~(Section \ref{sec:benchmark}). 
Given recent findings indicating that the CLIP models struggle to capture object information~\cite{DBLP:conf/emnlp/LiuJSWZ24}, spatial relationships~\cite{DBLP:conf/emnlp/KamathHC23a}, and compositional understanding~\cite{DBLP:conf/cvpr/ZhangAA24} in images during encoding, we hypothesize that these phenomena stem from the limited capabilities of the image encoders. Recent study~\cite{DBLP:journals/corr/abs-2408-04810} find that although scaling up the dataset and model parameters in the CLIP series proves effective for general tasks, such enhancements have minimal impact (even negative impact) on performance in complex tasks involving reasoning and relationship understanding. Therefore, we attempt to leverage the strengths of diverse types of encoders, which is also commonly used in general vqa tasks~\cite{DBLP:conf/cvpr/Tong0Z0LX24,DBLP:journals/corr/abs-2408-16357}, to enhance visual dependency and final performance.

SigLip~\cite{DBLP:conf/iccv/ZhaiM0B23} and DINO~\cite{DBLP:conf/iccv/CaronTMJMBJ21, DBLP:journals/tmlr/OquabDMVSKFHMEA24} are selected as the components of our combination, where SigLIP replaces the loss function used in CLIP with a simple pairwise sigmoid loss function, resulting in improved zero-shot classification performance; DINO, as a self-supervised visual model, is capable of capturing rich information from images. We consider two methods for feature concatenation:
\begin{itemize}[leftmargin=*]
    \item \textbf{Hidden Feature Concatenation} Images are input into multiple image encoders, and the dimensions of the hidden representation are concatenated, then mapped back to the original hidden layer output dimensions through the modality connection module. In this case, it is necessary for the image tokens across different representations to be consistent;
    \item \textbf{Image Token Concatenation} Images are input into multiple image encoders and mapped to the same hidden layer dimensions according to their respective  modality connection module, with the image tokens then concatenated. This method does not require the dimensions across different representations to be the same, but it increases the overall number of image tokens. \footnote{Experiment results of interleaving image tokens from different encoders as done in ~\cite{DBLP:conf/cvpr/Tong0Z0LX24} when combining multiple visual encoders can be found in Table~\ref{tab:combine_results}.}
\end{itemize}

\paragraph{Results} We experiment on G-LLaVA model~\footnote{Experiment results on MathLLavA model can be found in Table ~\ref{tab:mathllava_results}.}, and the results are shown in Table \ref{tab:image_encoder}.
We can find that:
\begin{itemize}[leftmargin=*]
    \item First, we find that integrating multiple image encoders, specifically SigLip and DINO, either separately or combined, significantly improves performance in general Visual Question Answering (VQA) tasks. The optimal performance is achieved by concatenating image tokens from all three encoders without dimensionality constraints, which elevates MME perception scores from 1481 to 1524.
    \item Second, an in-depth analysis of various metrics within the MME demonstrates notable advancements in color perception and spatial relationships (see Table~\ref{tab:detail_mme}). This suggests that utilizing diverse encoders allows the model to discern more detailed image features.
    \item Lastly, while enhancements are observed in general tasks, there is a consistent dip in performance on mathematical evaluation tasks, with only a slight increase seen in GeoQA. Moreover, the addition of multiple image encoders, such as combining SigLip and DINO, does not proportionally enhance performance beyond the use of either encoder on its own (i.e., the model's performance does not increase with the addition of more image encoders). For instance, in MathVista, the performance metrics for combining SigLip alone, DINO alone, and both combined are 30.8, 28.7, and 28.5, respectively. 
\end{itemize}

\paragraph{Hallucination Evaluation} 
We aim to further investigate the causes behind  this decline in performance. Given that the general improvement across common tasks suggests an enhanced general capability of the model, we are curious to determine if a more severe presence of hallucinations has led to this performance decline. Considering the lack of assessments for hallucinations in the mathematical domain, inspired by POPE~\cite{DBLP:conf/emnlp/LiDZWZW23}, we construct MathPOPE (9,000 questions based on 500 images) based on annotations of points in images from the Geometry3K~\cite{DBLP:conf/acl/LuGJQHLZ20} test set, in order to assess model hallucinations.  The
format of the question in MathPOPE is: `Is
there a point X in the image?’, where X refers to the
point name in picture. The questions in the dataset
are designed such that the objects are present and
absent in equal measure, therefore the ideal ‘yes’
response rate should be around 50\%.

Experimental results can be seen in Table \ref{tab:pope}. We observe that combining multiple image encoders indeed results in an increase in hallucinations to a certain extent. However, distinct from the general VQA tasks, in the domain of mathematical hallucination assessment, we note a reduction in hallucinations when three image encoders are combined (73.7 vs 81.0). Moreover, while there is a clear positive correlation between VQA performance in general tasks and hallucination performance, such correlation is absent in the domain of mathematical reasoning.

\paragraph{Analysis}
The results from the above experiments reveal a significant difference in the performance improvements within the mathematical domain compared to the findings in general VQA tasks. We believe there are two possible reasons for this. Firstly, images in the mathematical domain usually feature monochromatic colors and lower information density (see Figure \ref{fig:dataset}), which markedly distinguishes them from general images. Secondly, while there is a close association between performance and hallucination assessment in general tasks, the linkage between reasoning performance and hallucinations appears to be weaker. Therefore, reasoning tasks in the mathematical domain present unique challenges, making it significantly difficult to improve modality dependence. We leave further exploration to future work.

%% file: lvlm_eval.tex
\begin{table*}[t]
\setlength\tabcolsep{4pt}
\centering
\small
\resizebox{\linewidth}{!}{
\begin{tabular}{clcr@{\hspace{2pt}}lr@{\hspace{2pt}}lr@{\hspace{2pt}}lr@{\hspace{2pt}}lr@{\hspace{2pt}}lr@{\hspace{2pt}}l}
\toprule
\bf $\mathcal{D}_{\text{math}}$~\faCaretRight & & \textbf{Ins@Stage1/2/3} & \multicolumn{2}{c}{\textbf{GeoQA}} & \multicolumn{2}{c}{\textbf{${\text{Math}}$Verse}} & \multicolumn{2}{c}{\textbf{MathVista}} & \multicolumn{2}{c}{\textbf{MathVision}} & \multicolumn{2}{c}{\textbf{WeMath}} & \multicolumn{2}{c}{\textbf{Avg.}}\\
\midrule
\multirow{4}{*}{\bf \text{\faFileImageO~\faLongArrowRight~\faFileTextO}} & G-LLaVA & 618K/665K/117K & 68.0 & & 24.7 & & 34.9 & & 8.9 & & 39.3 & &35.2\\
& MathLLaVA& 558K/665K/339K &57.4 & & 25.6 & & \textbf{45.3} & & \textbf{9.5} & & 42.6 & &36.1\\
& MAVIS&1.1M/665K/555K & 70.6 & & \textbf{31.5} & & 37.0 & & 6.9 & & 39.4 & &37.1\\
& MultiMath& 1.2M/665K/1.0M &\textbf{74.4} & & 29.6 & & 44.5 & & 7.2 & & \textbf{44.7} & &\textbf{40.1}\\
\midrule
\multirow{4}{*}{\bf \text{\faFileImageO~\faRandom~\faFileTextO}}
& G-LLaVA & \multirow{4}{*}{\textit{same as above}} & 71.0 &$_{+3.0}$ & 22.4 &$_{-2.3}$ & \textbf{36.2} &$_{+1.9}$ & 8.2 &$_{-0.7}$ & 35.2 &$_{-4.1}$ &34.6&$_{-0.6}$\\
& MathLLaVA & & 56.6 &$_{-0.8}$ & 23.3 &$_{-2.3}$ & 34.4 &$_{-10.9}$ & 9.5 &$_{-0.0}$ & 38.5 &$_{-4.1}$ &32.5&$_{-3.6}$\\
& MAVIS & & 70.4 &$_{-0.2}$ & \textbf{26.0} &$_{-5.5}$ & 32.8 &$_{-4.2}$ & \textbf{10.2} &$_{+3.3}$ & 34.3 &$_{-5.1}$ &34.7&$_{-2.4}$\\
& MultiMath & & \textbf{76.5} &$_{+2.1}$ & 25.6 &$_{-4.0}$ & 35.0 &$_{-9.5}$ & 9.2 &$_{+2.0}$ & \textbf{39.5} &$_{-5.2}$ &\textbf{37.2}&$_{-2.9}$\\
\midrule
\multirow{4}{*}{\bf \text{\faFileO~\faLongArrowRight~\faFileTextO}}
& G-LLaVA &\multirow{4}{*}{\textit{same as above}} & 68.2 &$_{+0.2}$ & 23.7 &$_{-1.0}$ & 34.5 &$_{-0.4}$ & 8.6 &$_{-0.3}$ & \textbf{41.8} &$_{-2.5}$ &35.4&$_{+0.2}$\\
& MathLLaVA & & 56.9 &$_{-0.5}$ & 23.9 &$_{-1.7}$ & 36.2 &$_{-9.1}$ & 6.9 &$_{-2.6}$ & 39.7 &$_{-2.9}$ &32.7&$_{-3.4}$\\
& MAVIS & & 66.1 &$_{-4.5}$ & \textbf{27.0} &$_{-4.5}$ & 31.3 &$_{-5.7}$ & \textbf{10.2} &$_{+3.3}$ & 36.3 &$_{-3.1}$ &34.2&$_{-2.9}$\\
& MultiMath & & \textbf{70.3} &$_{-4.1}$ & 24.7 &$_{-4.9}$ & \textbf{37.1} &$_{-7.4}$ & 8.6 &$_{+1.2}$ & 41.1 &$_{-3.6}$ &\textbf{36.4}&$_{-3.7}$\\
\bottomrule
\toprule
\bf $\mathcal{D}_{\text{general}}$~\faCaretRight &  &\textbf{Ins@Stage1/2}& \multicolumn{2}{c}{\bf VQAv2} & \multicolumn{2}{c}{\bf MMBench} & \multicolumn{2}{c}{\bf MM-Vet} & \multicolumn{2}{c}{\bf MME\_P$^\dag$}  & \multicolumn{2}{c}{\bf MME\_C$^\dag$} & \multicolumn{2}{c}{\bf Avg.}\\
\midrule
\text{\faFileImageO~\faLongArrowRight~\faFileTextO} 
& LLaVA-1.5 & 558K/665K & \textbf{79.2}& &\textbf{66.8}& &\textbf{32.4}& &\textbf{1470.1}& &\textbf{322.1}& &\textbf{35.9}\\
\text{\faFileImageO~\faRandom~\faFileTextO}
& LLaVA-1.5 &\textit{same as above} & 46.2 &$_{-33.0}$ & 26.6 &$_{-40.2}$ & 12.2 &$_{-20.2}$ & 708.9 &$_{-761.2}$ & 276.8 &$_{-45.3}$ &17.1&$_{-18.8}$\\
\text{\faFileO~\faLongArrowRight~\faFileTextO}
& LLaVA-1.5 & \textit{same as above} & 60.8 &$_{-18.4}$ &54.3 &$_{-12.5}$ &20.9 &$_{-11.5}$& 706.0 &$_{-764.1}$ &271.8 &$_{-50.3}$ &27.3&$_{-8.6}$\\
\bottomrule
\end{tabular} }
\caption{Mathematical reasoning performance of using correct images, shuffled images, and no images during the training phase. 
``\textbf{Ins@Stage1/2/3}'' indicates the stages 1, 2, and 3 instances. The $\dag$ means we normalize MME\_P and MME\_C for the average score. }
\label{table:visual_modality}
\end{table*}

%% file: text_model.tex
\begin{table*}[htbp]
\centering
\small
\setlength\tabcolsep{4pt}
\begin{tabular}{lcccccccc}
\toprule
\textbf{Model}             & \textbf{Input} & \textbf{\#Param.} & \textbf{GeoQA} & \textbf{M.Verse} & \textbf{M.Vista} & \textbf{M.Vison} & \textbf{WeM.} & \textbf{Avg.} \\
\midrule
Deepseek-Math-RL~\citeyearpar{DBLP:journals/corr/abs-2402-03300}           & \faFileTextO       & 7B  & 39.8 & 20.7 & 26.1 & 11.5 & 33.3 &26.3\\
QWen-2.5-Math-Instruct~\citeyearpar{qwen2.5}     & \faFileTextO       & 7B  & 62.2 & 23.7 & 31.4 &   9.2  & 40.2 &33.3\\
Math-Shepherd-Mistral-RL~\citeyearpar{DBLP:conf/acl/WangLSXDLCWS24}   & \faFileTextO       & 7B  & 30.6 & 12.4 & 28.6 & 4.9  & 24.0 &20.1\\
OpenMath2-Llama3.1~\citeyearpar{DBLP:journals/corr/abs-2410-01560}         & \faFileTextO       & 8B  & 43.9 & 22.2 & 30.0 & 8.2  & 35.6 &28.0\\
Dart-Math-dsmath-prop2diff~\citeyearpar{DBLP:journals/corr/abs-2407-13690} & \faFileTextO       & 7B  & 43.6 & 19.2 & 26.6 & 11.5 & 38.1 & 27.8\\
Meta-Llama-3.1-Instruct~\citeyearpar{DBLP:journals/corr/abs-2407-21783}    & \faFileTextO       & 70B & 37.0 & 20.6 & 32.7 & 9.9  & 33.7 & 26.8\\
OpenMath2-Llama3.1 ~\citeyearpar{DBLP:journals/corr/abs-2410-01560}        & \faFileTextO       & 70B & 48.3 & 24.6 & 31.6 & 12.2 & \textbf{41.2} &31.6\\
Dart-Math-Llama3-prop2diff~\citeyearpar{DBLP:journals/corr/abs-2407-13690} & \faFileTextO       & 70B & 41.9 & 17.6 & 27.8 & 9.9  & 33.8 & 26.2 \\
QWen2.5-Math-Instruct~\citeyearpar{qwen2.5}      & \faFileTextO       & 72B & \textbf{68.0} & \textbf{32.3} & 34.6 & \textbf{12.5} & 32.9 & \textbf{36.1}\\
QWen2-Instruct-Step-DPO~\citeyearpar{DBLP:journals/corr/abs-2406-18629}    & \faFileTextO       & 72B & 61.9 & 21.8 & \textbf{35.0} & 9.9  & 30.9 & 31.9\\
GPT-4o~\citeyearpar{hurst2024gpt}                      & \faFileTextO       & /   & 57.7 & 25.4 & 30.7 & 10.5 & 33.9 & 31.6\\
\hline 
InternVL2~\citeyearpar{DBLP:journals/corr/abs-2312-14238}                & \faFileImageO+\faFileTextO & 8B  & 56.8 & 31.5 & 56.9 & 8.6  & 45.4 &39.8\\
QWen2-VL-Instruct~\citeyearpar{DBLP:journals/corr/abs-2409-12191}          & \faFileImageO+\faFileTextO & 8B  & 50.3 & 32.2 & 60.9 & 8.6  & 33.9 & 37.2\\
InternVL2-Llama3~\citeyearpar{DBLP:journals/corr/abs-2312-14238}           & \faFileImageO+\faFileTextO & 76B & 59.2 & 38.4 & 62.8 & 10.5 & 55.8 & 45.3\\
QWen2-VL-Instruct~\citeyearpar{DBLP:journals/corr/abs-2409-12191}          & \faFileImageO+\faFileTextO & 72B & \textbf{68.8} & 46.7 & \textbf{70.4} & 13.8 & 61.5 & \textbf{52.2}\\
GPT-4o~\citeyearpar{hurst2024gpt}                      & \faFileImageO+\faFileTextO & /   & 58.9 & \textbf{46.8} & 53.1 & \textbf{15.5} & \textbf{68.3} & 48.5\\
\bottomrule
\end{tabular}
\caption{Results of common text and visual language models on mathematical datasets. ``\textbf{\#Param.}'' indicates the number of parameters. ``\textbf{M.}'' is the short for ``Math''.}
\label{tab:text_model}
\end{table*}

%% file: appendix.tex
\section{Visual Control Experiment on HC-M3D Benchmark}

\begin{table}[t]
\setlength\tabcolsep{4pt}
\centering
\small
\begin{tabular}{clcccc}
\toprule
\bf $\mathcal{D}_{\text{math}}$~\faCaretRight &  & \textbf{ALL} $\uparrow$ & \textbf{DI} $\uparrow$ & \textbf{BC} $\uparrow$ & \textbf{AG} $\downarrow$ \\
\midrule
\multirow{2}{*}{\bf \text{\faFileImageO~\faLongArrowRight~\faFileTextO}} & G-LLaVA  & 45.4	& 41.5	& 15.2	& \bf 52.2    \\
& MathLLaVA &\bf 49.2	& \bf 44.8	& \bf 16.6	& 56.9\\
\midrule
\multirow{2}{*}{\bf \text{\faFileImageO~\faRandom~\faFileTextO}}
& G-LLaVA    & 41.2 & \bf 39.9	& \bf 12.1	& \bf 54.8    \\
& MultiMath  & \bf 42.3	& 38.8	& 11.2	& 60.6  \\

\midrule
\multirow{2}{*}{\bf \text{\faFileO~\faLongArrowRight~\faFileTextO}}
& G-LLaVA & 40.8	& 37.2	& \bf 10.8	& \bf 57.3    \\
& MultiMath  & \bf 42.8	& \bf 39.3	& 10.3	& 67.4   \\

\bottomrule
\end{tabular} 
\caption{Performance on HC-M3D datset of using correct images, shuffled images, and no images during the training phase. }
\label{table:visual_modality_hcm3d}
\end{table}

We evaluate whether HC-M3D can effectively distinguish the utilization of visual modalities on GLLaVA and MultiMath datasets. The results (shown in Table~\ref{table:visual_modality_hcm3d}) demonstrate that perturbing or removing the images leads to a significant decline in overall accuracy (ALL). Moreover, under the image removal setting, the prediction consistency (BC) is higher than in both the normal and perturbed image settings, suggesting that HC-M3D can partially identify the reliance on visual modalities.

\section{More Experiments on Exploring Different Visual Encoders}
\begin{table}[!htbp]
\centering
\resizebox{\linewidth}{!}{
\begin{tabular}{c|ccccc}
\toprule
& \rotatebox[origin=l]{90}{\textbf{GeoQA}}
 & \rotatebox[origin=l]{90}{\textbf{MathVerse}}& \bf\rotatebox[origin=l]{90}{MathVista}	& \bf \rotatebox[origin=l]{90}{MathVision}	& \bf \rotatebox[origin=l]{90}{WeMath} \\
\midrule
clip-L          &57.4&\bf 25.6	& \bf 45.3	& 9.5	& \bf 42.6\\
clip-L+siglip-L &57.3&  24.3	& 44.0	& 10.9	& 40.6 \\
clip-L+dinov2-L &\bf 58.6 & 24.0	& 39.6	& \bf 12.2	& 38.8  \\
clip-L+siglip-L+dinov2-L &58.0& 24.5 & 37.6	& 8.2	&37.9\\      
\bottomrule
\end{tabular}
}
\caption{Results of combining different visual encoders on mathematical tasks. Experiments are conducted on MathLLaVA model. }
\label{tab:mathllava_results}
\end{table}

we additionally report results on the training set of MathLLaVA to provide a more comprehensive evaluation. (as shown in Table ~\ref{table:visual_modality}, MathLLaVA demonstrates a stronger dependency on image-based information. )
Here we combine different encoders by concatenating image tokens.
The experimental results on MathLLaVA (shown in Table~\ref{tab:mathllava_results}) are generally consistent with those on G-LLaVA (Table ~\ref{tab:image_encoder}). Combining multiple encoders does not enhance the performance of complex reasoning on images. 
Although Math-LLaVA's dataset exhibits a marginally higher degree of visual-textual dependency, the 3\% disparity is not deemed to constitute a robust reliance (e.g., when compared with the LLaVA's training dataset). Consequently, it is not feasible to derive a definitive conclusion from such a dataset. The development of an additional general geometry dataset characterized by a pronounced visual-textual dependency is reserved for future research endeavors.


\begin{table}[!htbp]
\centering
\resizebox{\linewidth}{!}{
\begin{tabular}{c|cccccc|c}
\toprule
& \rotatebox[origin=l]{90}{\textbf{GeoQA}}
 & \rotatebox[origin=l]{90}{\textbf{MathVerse}}& \bf\rotatebox[origin=l]{90}{MathVista}	& \bf \rotatebox[origin=l]{90}{MathVision}	& \bf \rotatebox[origin=l]{90}{WeMath} & \bf \rotatebox[origin=l]{90}{HC-M3D} & \bf \rotatebox[origin=l]{90}{Avg.}\\
 
\midrule
clip-L          &68.0 &\bf  24.7 & \bf 34.9 & \bf 8.9& \bf 39.3 & \bf 45.4 & \bf 36.9\\
clip-L+dinov2-L &\bf 70.7 & 22.3 & 28.7 & 7.6& 35.7 & 39.2 & 34.0  \\
clip-L+dinov2-L~(\citeyear{DBLP:conf/cvpr/Tong0Z0LX24}) &66.7 & 24.3 & 31.9 & 6.6& 38.7 & 41.8 & 35.0  \\
\bottomrule
\end{tabular}
}
\caption{Results of interleaving the images tokens from different encoders as done in ~\cite{DBLP:conf/cvpr/Tong0Z0LX24}.}
\label{tab:combine_results}
\end{table}

We also try interleaving the tokens from different encoders as done in ~\cite{DBLP:conf/cvpr/Tong0Z0LX24} when combining multiple visual encoders.
Experimental results (shown in Table ~\ref{tab:combine_results}) indicate that the interleaving input strategy achieved slightly better performance (average score of 35.0) compared to directly concatenating the representations of CLIP and DINOv2 (average score of 34.0). However, the performance of this method remains significantly lower than that of using CLIP representations alone (average score of 36.9).

\section{Training Details}

\begin{table*}[t]
 \centering
\resizebox{0.85\linewidth}{!}{
\begin{tabular}{ccccc}
\toprule
\textbf{Statistcs} & \textbf{G-LLaVA} & \textbf{MathLLaVA} & \textbf{MAVIS} & \textbf{MultiMath} \\
\midrule
\multicolumn{5}{c}{\bf Pretrain stage}  \\
\hline
Dataset            & \begin{tabular}[c]{@{}c@{}}Geo170K-align\\ +LLaVA-Pretrain\end{tabular} & LLaVA-Pretrain     & \begin{tabular}[c]{@{}c@{}}MAVIS-align\\ +LLava-Pretrain\end{tabular} & \begin{tabular}[c]{@{}c@{}}MultiMath-300K-align\\ +Geo170K-align\\ +LLaVA-Pretrain\end{tabular} \\
\#Samples & 618K & 558K & 1.1M & 1.2M \\
Training Module & mm adapter & mm adapter & mm adapter & mm adapter \\
Epoch & 1 & 1 & 1& 1\\
Batch\_size * \#gpu &32*8&32*8&32*8&32*8\\
Learning Rate & 1e-3& 1e-3& 1e-3& 1e-3\\
\hline
\multicolumn{5}{c}{\bf SFT stage} \\
\hline
Dataset & LLaVA-Instruction & LLaVA-Instruction  & LLaVA-Instruction & LLaVA-Instruction \\
\#Samples & 665K & 665K & 665K & 665K \\
Training Module    & mm adapter+llm & mm adapter+llm & mm adapter+llm & mm adapter+llm \\
Epoch & 1 & 1 & 1& 1\\
Batch\_size * \#gpu &8*8&8*8&8*8&8*8\\
Learning Rate & 2e-5& 2e-5& 2e-5& 2e-5\\
\hline
\multicolumn{5}{c}{\bf Math SFT stage} \\
\hline
Dataset & Geo170K-qa & MathV360K & MAVIS-qa & \begin{tabular}[c]{@{}c@{}}MultiMath-300K-qa\\ +Geo170K-qa\\ +MathV360K\end{tabular}    \\
\#Samples & 117K & 339K & 555K & 1.0M \\
Training Module    & mm adapter+llm & mm adapter+llm & mm adapter+llm & mm adapter+llm \\     
Epoch & 2 & 2 & 2& 2\\
Batch\_size * \#gpu &8*8&8*8&8*8&8*8\\
Learning Rate & 2e-5& 2e-5& 2e-5& 2e-5\\
\bottomrule
\end{tabular}
}
\caption{Training detail information across different models.}
\label{tab:training_details}
\end{table*}


Detailed information on model training are shown in \ref{tab:training_details}. Given that existing multimodal mathematical models are based on the same LLaVA architecture with the main differentiation being the training data, we standardize the training approach as three stage: Pretrain, SFT and Math SFT with language model \texttt{deepseek-math-rl-7b}. All the models are trained
on 8 NVIDIA A100-80GB GPUs with the random seed 42.

\begin{table*}[t]
\centering
\resizebox{\linewidth}{!}{
\begin{tabular}{ccccccccccccccc}
\toprule
\textbf{Image Encoder}    & \textbf{existence} & \textbf{count} & \textbf{position} & \textbf{color} & \textbf{posters} & \textbf{celebrity} & \textbf{scene} & \textbf{landmark} & \textbf{artwork} & \textbf{OCR}   & \textbf{CR} & \textbf{NC} & \textbf{TT} & \textbf{CR} \\
\midrule
clip-B                    & 185.0              & 141.7          & 141.7             & 150.0          & 120.7            & 105.0              & \textbf{163.3} & 138.5             & 102.5            & 117.5          & 108.6                                                                    & 62.5                                                                     & 50.0                                                                & 52.5                                                              \\
clip-L                    & 195.0              & 156.7          & 131.7             & 175.0          & 136.7            & 127.6              & 160.5          & 147.0             & 118.3            & 132.5          & \textbf{112.1}                                                           & 47.5                                                                     & 50.0                                                                & 60.0                                                              \\
\hline
\multicolumn{15}{c}{(a) combine features by concatenating the hidden feature}                                                                                                                                                                                                                                                                                                                                                                                                                                         \\
\hline
clip-B+ siglip-B          & \textbf{200.0}     & 151.7          & 140.0             & 160.0          & 114.6            & 113.2              & 158.5          & 140.3             & 111.8            & 107.5          & 110.0                                                                    & 47.5                                                                     & 50.0                                                                & 50.0                                                              \\
clip-L+dinov2-L           & 195.0              & 148.3          & \textbf{143.3}    & \textbf{195.0} & 131.3            & 122.6              & 158.0          & 147.3             & 119.3            & 140.0          & 105.7                                                                    & 47.5                                                                     & 50.0                                                                & 50.0                                                              \\
clip-B+siglip-B, dino-B   & 195.0              & 151.7          & 136.7             & 175.0          & 107.8            & 108.8              & 158.3          & 140.5             & 105.3            & 125.0          & 110.0                                                                    & 65.0                                                                     & 50.0                                                                & 55.0                                                              \\
\hline
\multicolumn{15}{c}{(b) combine features by increasing image token number}                                                                                                                                                                                                                                                                                                                                                                                                                                            \\
\hline
clip-L+siglip-L           & 190.0              & 145.0          & 120.0             & 185.0          & 117.3            & 120.6              & 159.3          & 158.0             & 117.5            & 132.5          & 101.4                                                                    & 42.5                                                                     & \textbf{55.0}                                                       & \textbf{67.5}                                                     \\
clip-L+dinov2-L           & \textbf{200.0}     & 158.3          & 141.7             & 180.0          & 136.1            & 117.6              & 160.5          & 155.3             & \textbf{122.8}   & 122.5          & 107.1                                                                    & 57.5                                                                     & 50.0                                                                & 55.0                                                              \\
clip-L+siglip-L, dinov2-L & 185.0              & \textbf{160.0} & 130.0             & 178.3          & \textbf{137.4}   & \textbf{137.1}     & 161.3          & \textbf{165.3}    & \textbf{122.8}   & \textbf{147.5} & 109.3                                                                    & \textbf{80.0}                                                            & 50.0                                                                & 62.5    \\       
\bottomrule
\end{tabular}
}
\caption{Detailed results on the MME dataset (Supplementary to Table \ref{tab:image_encoder}). In the table, CR, NC, TT, and CR correspond to commonsense reasoning, numerical calculation, text translation, and code reasoning, respectively.}
\label{tab:detail_mme}
\end{table*}

\begin{table*}[t]
\centering
\resizebox{0.9\linewidth}{!}{
\begin{tabular}{cccccccccccc}
\toprule
\multicolumn{1}{l}{\multirow{2}{*}{\textbf{Image Encoder}}} & \multicolumn{5}{c}{\textbf{MathVista}}                                                                                                  & \multicolumn{3}{c}{\textbf{MathVerse}}                                                                                                                       & \multicolumn{3}{c}{\textbf{HC-M3D}}          \\
\cmidrule(r){2-6} \cmidrule(r){7-9} \cmidrule(r){10-12}
\multicolumn{1}{l}{}                                        & \textbf{TextQA} & \textbf{VQA}  & \textbf{geometry} & \textbf{\begin{tabular}[c]{@{}c@{}}Math \\ Word\end{tabular}} & \textbf{FigureQA} & \textbf{\begin{tabular}[c]{@{}c@{}}Plane \\ Geometry\end{tabular}} & \textbf{Functions} & \textbf{\begin{tabular}[c]{@{}c@{}}Solid \\ Geometry\end{tabular}} & \textbf{DI}   & \textbf{BC}   & \textbf{AG}   \\
\midrule

clip-B                    & 33.5            & 27.4          & 62.0              & 17.2                                                          & 19.0              & 29.7                                                               & 17.1               & 23.0                                                               & 39.6          & 8.6           & 62.0 \\
clip-L                    & \textbf{38.6}   & \textbf{35.2} & 62.0              & \textbf{21.0}                                                 & 21.2              & \textbf{31.4}                                                      & \textbf{19.1}      & \textbf{24.7}                                                      & \textbf{40.8} & \textbf{14.5} & \textbf{53.4}          \\
\hline
\multicolumn{12}{c}{(a) combine features by concatenating the hidden feature}                                                                                                                                                                                                                                                                                                      \\
\hline
clip-B+ siglip-B          & 37.3            & 24.0          & \textbf{64.4}     & 12.9                                                          & 19.0              & 27.8                                                               & 18.1               & 22.2                                                               & 37.5          & 11.0          & 56.6          \\
clip-L+dinov2-L           & 36.1            & 25.7          & 59.1              & 11.3                                                          & 19.7              & 29.8                                                               & 18.1               & 23.2                                                               & 38.8          & 12.1          & 60.8          \\
clip-B+siglip-B, dino-B   & 38.6            & 26.8          & 60.6              & 12.4                                                          & 19.3              & 28.8                                                               & 17.5               & 22.5                                                               & 38.3          & 9.8           & 58.3          \\
\hline
\multicolumn{12}{c}{(b) combine features by increasing image token number}                                                                                                                                                                                                                                                                                                         \\
\hline
clip-L+siglip-L           & 31.7            & 31.8          & 54.8              & 14.0                                                          & \textbf{22.7}     & 29.4                                                               & 16.9               & 22.6                                                               & 37.7          & 11.0          & 59.4          \\
clip-L+dinov2-L           & 32.3            & 24.0          & 53.9              & 15.1                                                          & 19.7              & 29.4                                                               & 15.7               & 22.3                                                               & 37.3          & 9.8           & 55.0          \\
clip-L+siglip-L, dinov2-L & 29.1            & 24.6          & 58.7              & 11.8                                                          & 19.0              & 29.4                                                               & 16.9               & 22.6                                                               & 36.8          & 9.6           & 57.1  \\
\bottomrule
\end{tabular}
}
\caption{Detailed results on MathVista, MathVerse, and HC-M3D (Supplement to Table \ref{tab:image_encoder}). Except for the AG (agreement) metric (lower is better), all other metrics indicate higher is better.}
\label{tab:detail_math}
\end{table*}

\begin{figure*}[htbp]
    \centering
    \includegraphics[width=0.95\textwidth]{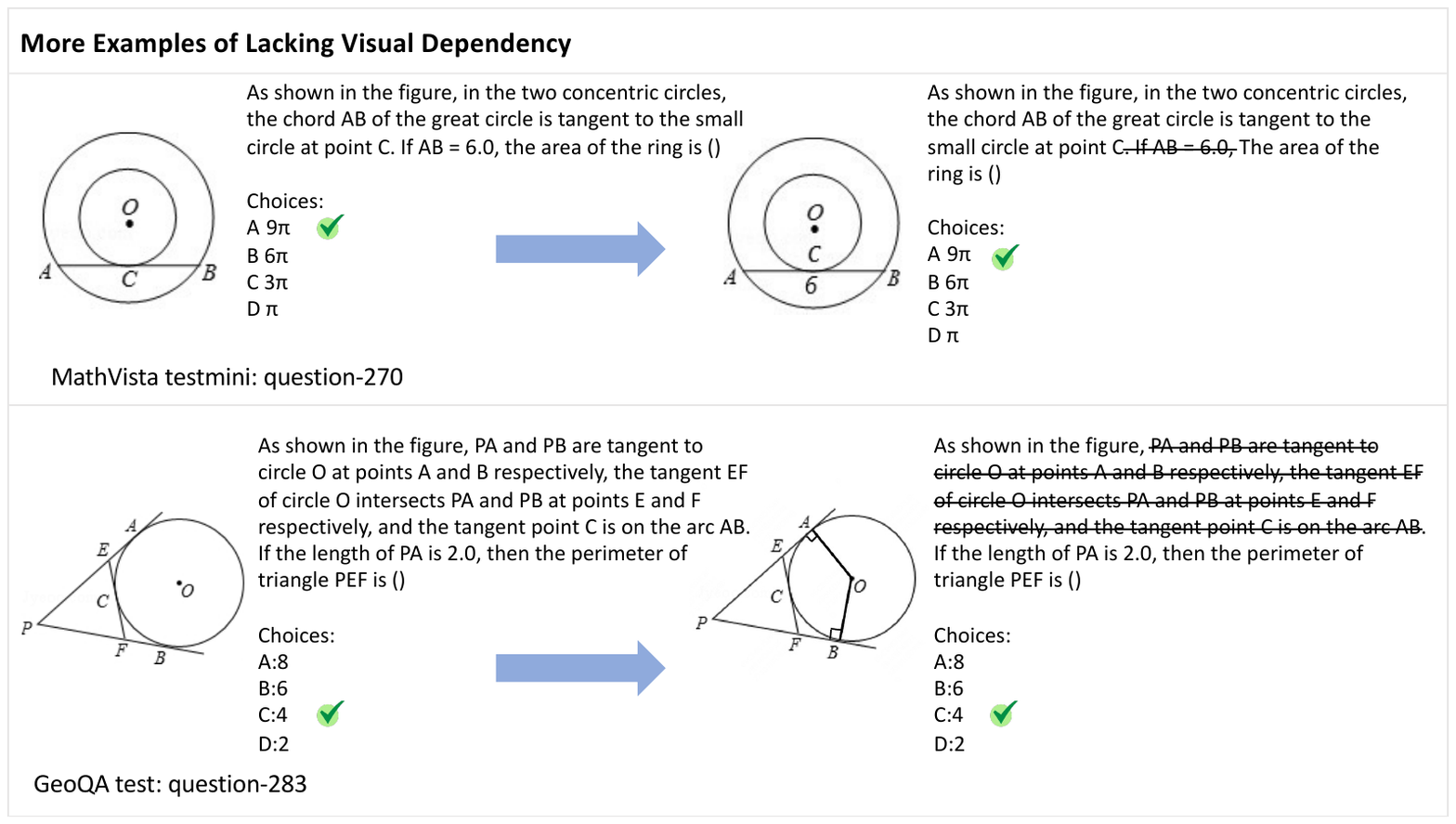}
    \caption{More examples from our constructed HC-M3D benchmark where original samples are lacking visual dependency. (Supplement to Figure \ref{fig:dataset})}
    \label{fig:dependency}
\end{figure*}

\begin{figure*}[htbp]
    \centering
    \includegraphics[width=0.95\textwidth]{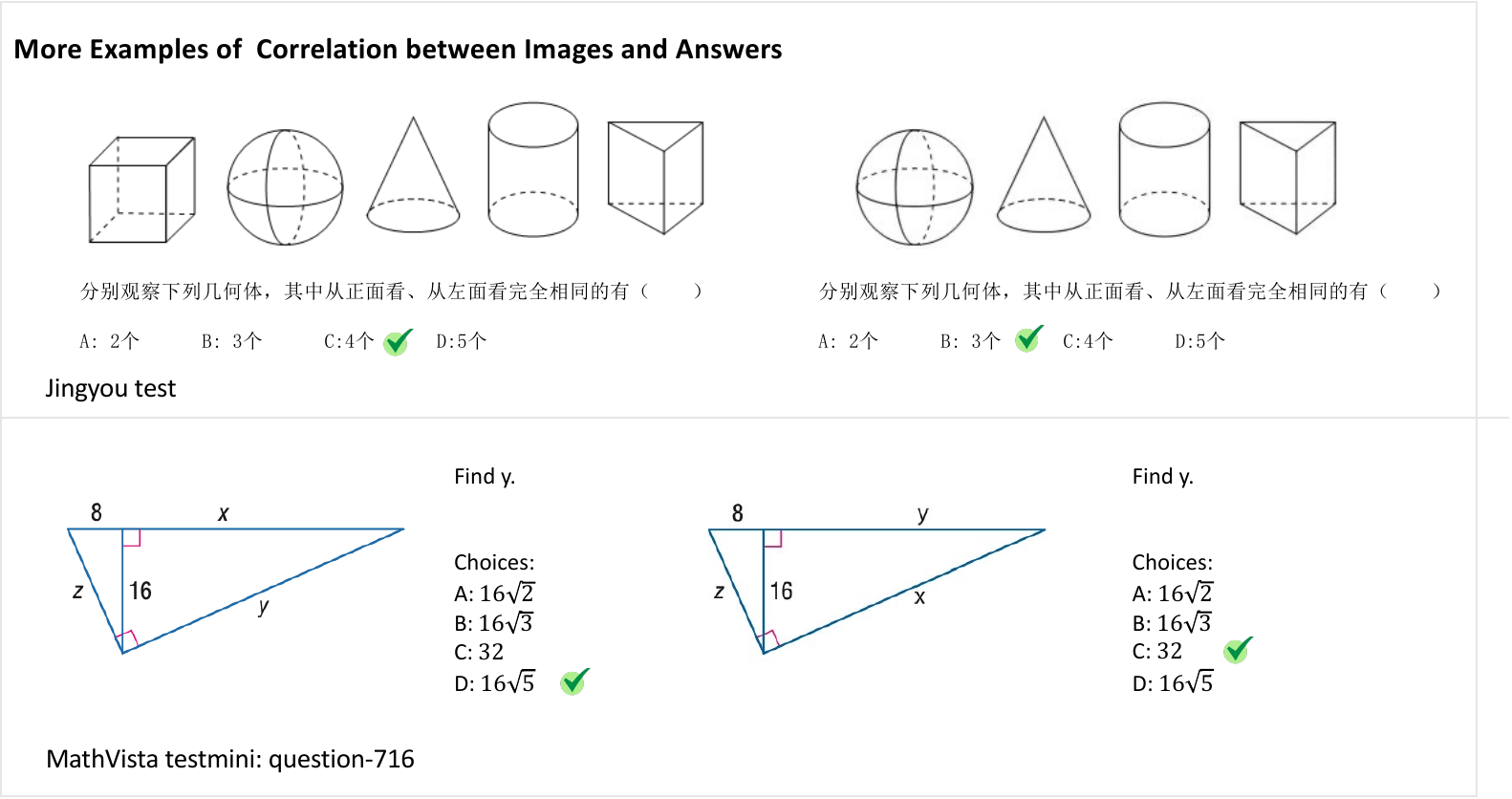}
    \caption{More examples from our constructed HC-M3D benchmark where images highly related to answers. (Supplement to Figure \ref{fig:dataset})}
    \label{fig:connection}
\end{figure*}

\clearpage

\begin{figure*}[htbp]
    \centering
    \includegraphics[width=0.95\textwidth]{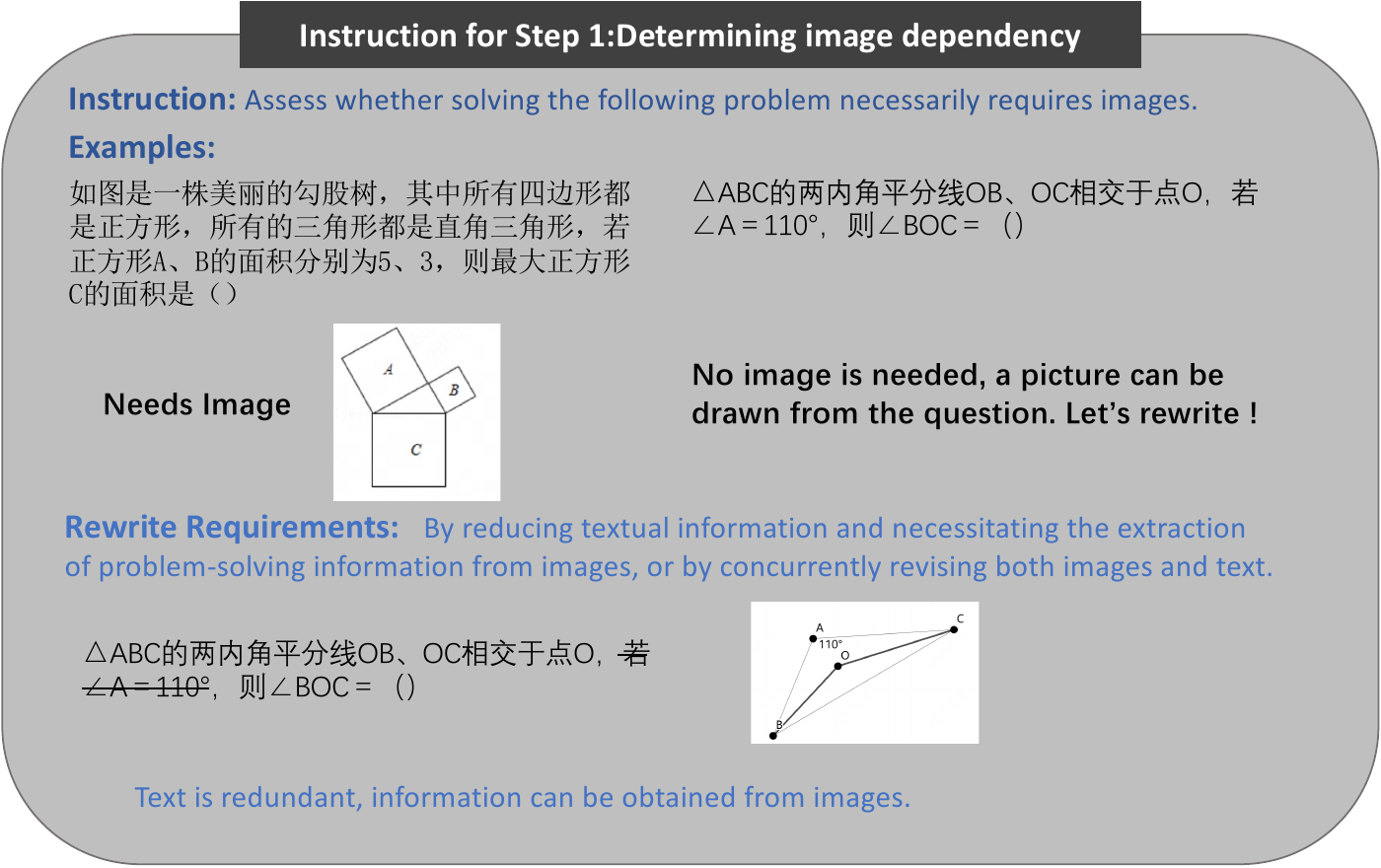}
    \caption{Instruction for step 1: determining image dependency.}
    \label{fig:instruction1}
\end{figure*}

\begin{figure*}[htbp]
    \centering
    \includegraphics[width=0.95\textwidth]{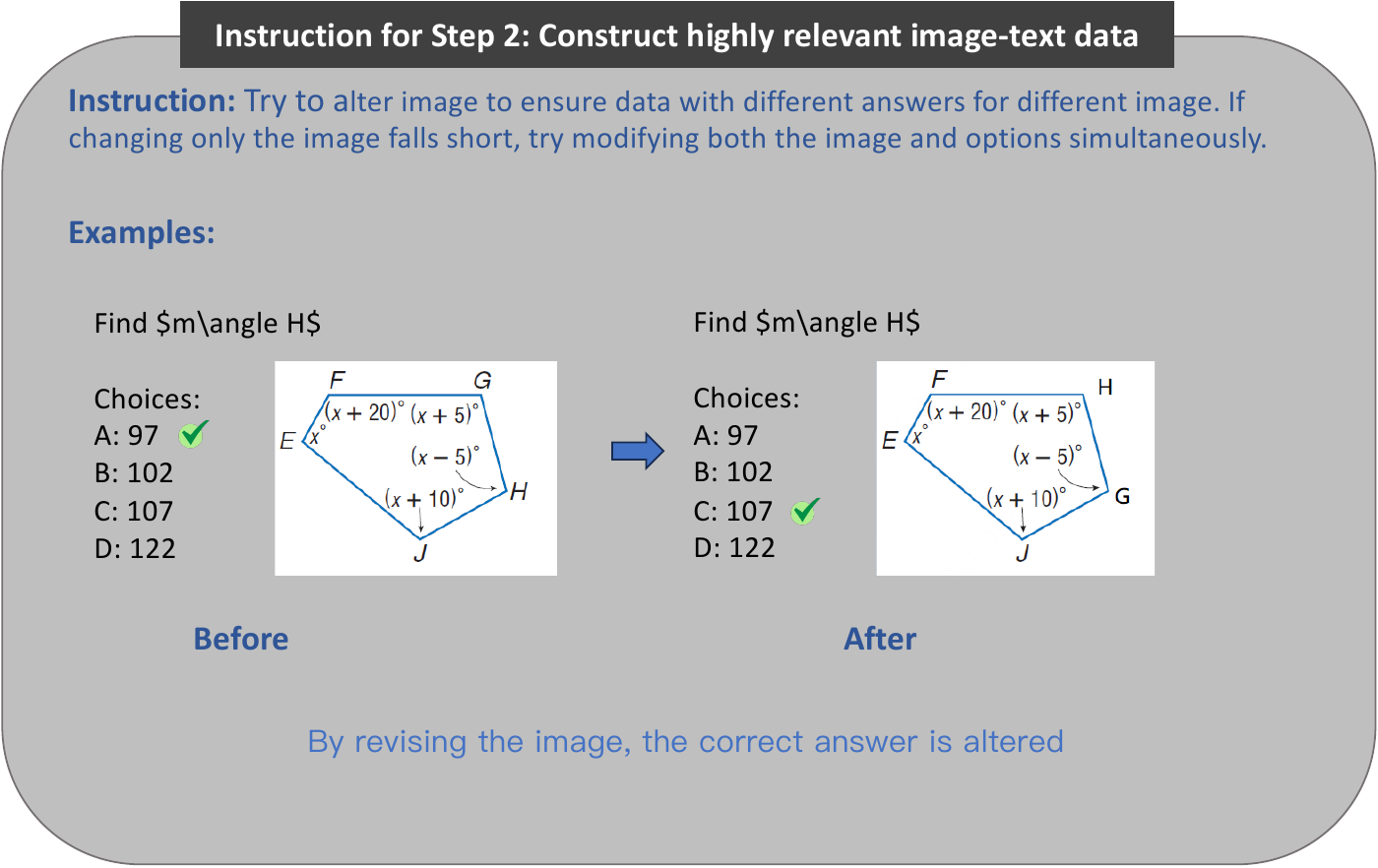}
    \caption{Instruction for step 2: construct highly relevant image-text data.}
    \label{fig:instruction2}
\end{figure*}